\documentclass[11pt,a4paper]{article}
\usepackage{times,latexsym}
\usepackage{url}
\usepackage[T1]{fontenc}
\usepackage[table]{xcolor}
\definecolor{myblue}{RGB}{87, 183, 242}
\definecolor{myyellow}{RGB}{244, 166, 42}
\definecolor{mygrey}{RGB}{200,200,200}

\usepackage[acceptedWithA]{tacl2021v1}

\usepackage{booktabs}
\usepackage{tcolorbox}
\usepackage{expex}
\usepackage{amsmath}

\usepackage{xspace,mfirstuc,tabulary}

\newif\iftaclinstructions
\taclinstructionsfalse 
\iftaclinstructions
\renewcommand{\confidential}{}
\renewcommand{\anonsubtext}{(No author info supplied here, for consistency with
TACL-submission anonymization requirements)}
\newcommand{\instr}
\fi

\iftaclpubformat 

\else

\fi

\title{The Structural Sources of Verb Meaning Revisited: \\ Large Language Models Display Syntactic Bootstrapping}

\author{Xiaomeng Zhu
\hspace{0.7in} R. Thomas McCoy  \hspace{0.7in} Robert Frank\\ Department of Linguistics, Yale University\\Wu Tsai Institute, Yale University\\ \texttt{\{miranda.zhu, tom.mccoy, robert.frank\}@yale.edu}}

\date{}

\begin{document}
\maketitle

\begin{figure*}[ht]
    \centering
    \includegraphics[width=1\linewidth]{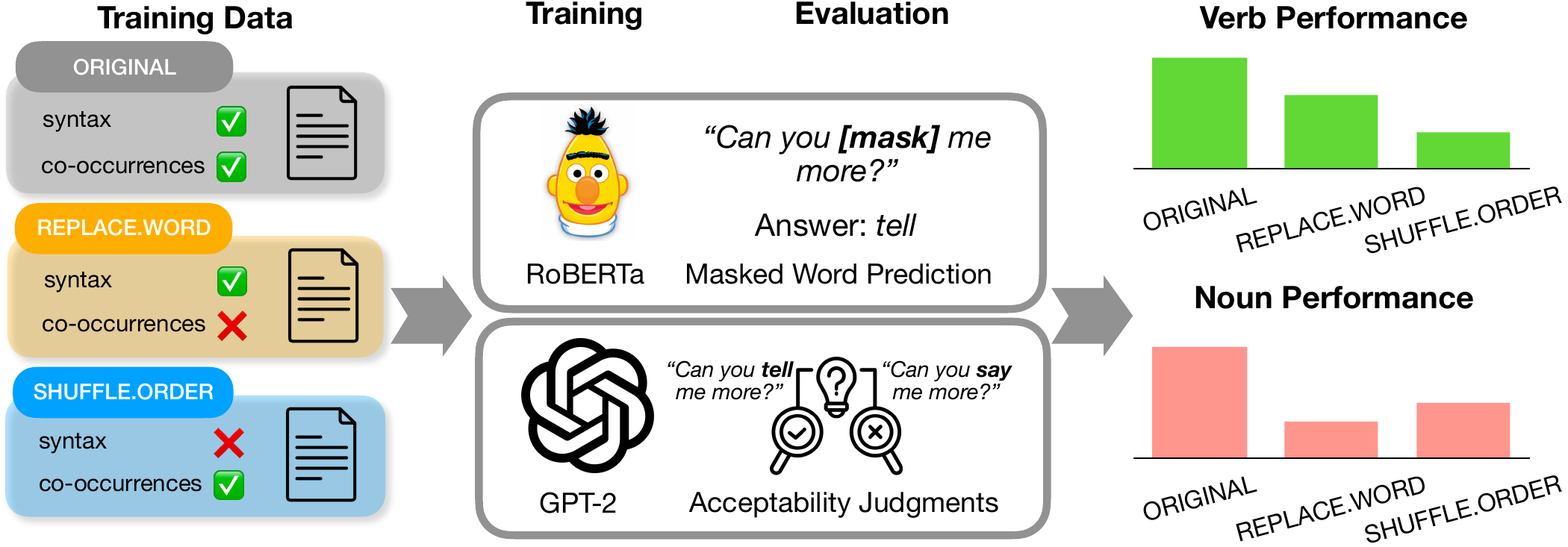}
    \caption{Illustration of our experimental setup and results.}
    \label{fig:overall}
\end{figure*}

\begin{abstract}
  Syntactic bootstrapping \citep{gleitman1990structural} is the hypothesis that children use the syntactic environments in which a verb occurs to learn its meaning. In this paper, we examine whether large language models exhibit a similar behavior. We do this by training RoBERTa and GPT-2 on perturbed datasets where syntactic information is ablated. 
  Our results show that models' verb representation degrades more when syntactic cues are removed than when co-occurrence information is removed.
  Furthermore, the representation of mental verbs, for which syntactic bootstrapping has been shown to be particularly crucial in human verb learning, is more negatively impacted in such training regimes than physical verbs. In contrast, models' representation of nouns is affected more when co-occurrences are distorted than when syntax is distorted. In addition to reinforcing the important role of syntactic bootstrapping in verb learning, our results demonstrated the viability of testing developmental hypotheses on a larger scale through manipulating the learning environments of large language models.
\end{abstract}

\section{Introduction}

One of the most important questions in language acquisition is how children acquire word meanings \citep{carey1978child, chomsky1993language, gelman1998beyond, bloom2002children, waxman2006early}. An intuitive answer is to say that they do so by observing word-world co-occurrences, which is learning by keeping track of which words co-occur with which situations \citep{locke1847essay}.  While this account might work for nouns, it presents an interesting puzzle for verbs \citep{gleitman1990structural}: the events in the world that correspond to verb meanings are often not observable at all.
For example, sentence (\ref{ex:see}) with \textit{see} could occur when the utterer is about to make a phone call or when they are actively looking at something. 
\ex[aboveexskip=7pt, belowexskip=7pt]<unobseravable> \label{ex:see}
Let's see if Granny's home!
\xe
This makes cross-situational contingencies an unreliable cue for predicting when the verb occurs. This observation underlies the proposal of \textbf{syntactic bootstrapping} \citep{gleitman1990structural}, which argues that in addition to word-world contingencies, the syntactic environments in which a verb occurs serve as another important source of information in its acquisition. 
For instance, the fact that \textit{see} can take a clausal complement, such as \textit{if Granny's home} in (\ref{ex:see}), provides a clue that \textit{see} can be a mental verb that indicates something about the status of a proposition in a person's mind.

In this paper, we explore whether syntactic bootstrapping also occurs in large language models (LLMs). We examine the relative contribution of syntactic cues (i.e., word order) vs.\ distributional co-occurrences (i.e., other words that co-occur with the verb) on the learned verb representations. To do this, we intervene on the training data in ways that remove information about syntax or co-occurrence and test how these ablations affect models' performance on targeted evaluation sets. We find that verb learning is more negatively impacted by ablating syntactic information than by ablating co-occurrence information, while the pattern is the reverse for nouns (see Figure \ref{fig:overall}). This suggests that syntactic information does indeed play a significant role in the learning of verb meanings by LLMs. The fact that machines, like humans, display syntactic bootstrapping suggests not only that structural cues are important for verbs (as opposed to co-occurrences for nouns) but also that humans and models can both capitalize on such statistical correlations as a common strategy for word learning.

\section{Word Learning} \label{sec:synboo}
How do children acquire word meanings? This has been a long-standing problem in the field of language acquisition. One appealing account is that words are learned through observing word-world contingencies. For example, if the word \textit{cat} most often occurs in scenes that involve a small domesticated feline animal, then children might make the inference that \textit{cat} refers to this type of object. Although the full picture is much more complicated with more problems that need to be addressed (e.g., the induction problem of how children arrive at the precise meaning of cat instead of its superset meaning such as ``mammal'' or ``animal'' that is fully compatible with the scene \citep{quine1960word}), it is generally agreed upon that word-to-world mapping is a crucial information source for learning the first set of nouns \citep{snedeker2000cross}. 

Verbs, however, pose a challenge to such observational accounts of word learning \citep{gleitman1990structural}. For example, verbs are often used in contexts in which the corresponding event is not taking place. Furthermore, mental verbs such as \textit{think}, \textit{guess}, \textit{wonder}, and \textit{know} do not refer to the observable world at all, and some verb pairs can be used to describe the same event (e.g. \textit{buy}-\textit{sell} and \textit{chase}-\textit{flee}). In these cases, the word-world contingencies are simply not sufficient for verb meaning acquisition. In response to this problem, \citeauthor{gleitman1990structural}\ proposed that \textbf{syntactic bootstrapping} plays a key role in acquisition: children not only observe word-world co-occurrences but also use the syntactic environments in which a verb occurs to infer its meaning. 
For example, given the sentence frame of (2a), a learner might infer that verb1 encodes some causative action where Abigail is the doer and Betty is the undergoer. 
\pex[aboveexskip=3pt, belowexskip=3pt]<verbmeaning>
\a Abigail [verb1] Betty.
\a Abigail and Betty [verb2].
\xe 
In contrast, it is less possible for verb2 in (2b) to encode the same causative meaning because Abigail and Betty are now both in the subject position that precedes the verb. 

A rich body of experimental work has subsequently provided concrete support for syntactic bootstrapping (see \citet{babineau2024syntactic} for a comprehensive review). For example,
\citet{naigles1990children} found that children with ages between 1 year 11 months and 2 years 3 months were able to associate novel verbs that appear in transitive frames like (\getref{verbmeaning}a) with causative actions (such as an individual pushing another into a squatting position) and intransitive frames like (\getref{verbmeaning}b) with non-causative actions (such as an individual waving their arm). Such associations persist even under the absence of referential events during the first exposure to novel verbs, with the number of nouns being a simple but important structural cue for verb meaning \citep{yuan2009really, yuan2012counting}.

The special role of syntactic structure in inferring verb meaning has also been shown for adult speakers. \citet{gillette1999human} developed the \textbf{Human Simulation Paradigm}, where adult participants were recruited to guess the common word that is produced by mothers across each of 6 muted videos showing them playing with their children. Results showed that the identification accuracy for nouns (45\%) is significantly higher than for verbs (15\%). More notably, one-third of the verbs were never correctly identified, which underscores the difficulty of identifying verbs through cross-situational observations.
In terms of verbs, \citeauthor{gillette1999human}\ found that the identification accuracy was much higher in a condition where only syntactic information was provided (51.7\%) than in other conditions where cross-situational videos or co-occurring nouns were provided. 
Furthermore, the verbs that are hard to identify given videos (namely mental verbs like \textit{see}, \textit{look}, and \textit{want}) became the ones that were most easily identified when participants were provided with syntactic frames alone.

Despite the empirical support for people's reliance on syntactic structure when making inferences about verbs, \citet{gillette1999human} argued that observational learning is still the foundation of word learning: a small set of concrete nouns is learned first via observing word-world co-occurrences, which is also consistent with the finding that nouns dominate children's early vocabulary \citep{bates1996individual}. This set of learned nouns enables the learning of clause-level syntax, which paves the way for subsequent verb learning. The number of arguments that a predicate takes can then be inferred from the number of nouns, which helps to build the phrase structure that bootstraps verb learning.

To sum up, although the Human Simulation Paradigm did not provide direct evidence for the employment of syntactic structure during verb learning or the learning mechanism itself, it successfully reveals what kinds of information are needed to identify words and how nouns and verbs differ: \textbf{syntactic structure} is more helpful than \textbf{co-occurrence information} in predicting verb identities, while nouns can be much more easily identified via cross-situational observations. Although both sources of information interact with each other during the word acquisition process, we focus on the difference in relative strength between these two for verbs vs.\ nouns, respectively. We propose that syntactic bootstrapping can serve as an interesting window for understanding how LLMs acquire meanings, and conversely, as artificial learners, LLMs can also help to examine hypotheses in the language acquisition literature.

\section{Problems of Acquisition: Insights from LLMs} \label{sec:comp}

The advancement of neural language models has provided new ways for examining puzzles of language acquisition. As weakly-biased artificial learners of human language, they can help us investigate the necessity of strong innate predispositions that have been proposed to address these puzzles (e.g. \citealp{chomsky1980rules, mccoytacl, warstadt2020neuralnetworksacquirestructural, yedetore-etal-2023-poor, murty-etal-2023-grokking}). 

In this paper, we maintain the same computational perspective but instead focus on the problem of verb learning, specifically the syntactic bootstrapping hypothesis that is outlined in the previous section. We aim to provide a computational parallel of the Human Simulation Paradigm: \citeauthor{gillette1999human}\ successfully demonstrated the distinction in information sources that are most helpful for identifying words of different categories: cross-situational occurrences for nouns vs.\ syntactic structure for verbs. However, the role of this association \textit{during} the acquisition process remains obscure, given that the experimental paradigm only involved adult participants. Meanwhile, experiments with children \citep[e.g.,][]{naigles1990children} have been valuable for showing that children can, in principle, leverage syntactic bootstrapping, but it is unclear whether syntactic information plays the same role in learning from naturalistic text as it does in learning from controlled experimental stimuli. 
Are learners actually able to capitalize on this distinction for acquiring meanings from naturalistic input? 

The most direct way of answering this question would be to remove these sources of information during learning and then compare the linguistic knowledge gained by the learner in respective cases. This design is only possible in the case of artificial learners such as LLMs: the training data can be systematically manipulated to reveal dependence of linguistic constructions during acquisition \citep{leong2024testing, patil-etal-2024-filtered, misra-mahowald-2024-language}. We therefore analyze the acquisition of word meanings in LLMs to gain a window into which types of information in naturalistic linguistic input are most useful for statistical learning. Specifically, we identify the following research questions:
\begin{itemize}
    \item RQ1: Do LLMs use syntactic information to learn verb meanings? 
    \item RQ2: If so, do mental and physical verbs differ in terms of their reliance on syntax?
    \item RQ3: Does noun learning exhibit the same pattern as verb learning in LLMs?
\end{itemize}

To the best of our knowledge, no previous studies have directly examined these questions. Most of the previous literature on LLM meaning acquisition either focused on the role of distributional information \citep{mikolov2013efficientestimationwordrepresentations} or nouns \citep{huebner2018structured, huebner2020order, huebner2023analogical} specifically, without examining the importance of syntactic information. While \citet{sinha-etal-2021-masked} did study the effect of word order on downstream natural language understanding tasks (e.g., classifying whether a sentence has a positive or negative sentiment), it did not examine its effect on verb meaning specifically.
This is the exact gap we aim to address.
As a preview of our results, we found that syntactic information plays a more important role in verb meaning acquisition than co-occurrence information, especially for mental verbs (Section \ref{sec:exp1}), whereas co-occurrence information is more important than syntactic information for noun learning (Section \ref{sec:exp2}). We discuss the implications of our results in Section \ref{sec:discussion}.

\section{Methods} \label{sec:methods}

We consider three versions of the training set: the original dataset, a perturbation that ablates syntax, and a perturbation that ablates co-occurrences for verbs. We train LLMs in these three settings while holding constant the architectures and training algorithm. We then evaluate models on a targeted evaluation set for verb meaning. The respective role of syntax vs.\ co-occurrences can then be revealed through performance differences.\footnote{All model checkpoints, dataset, and results will be made publicly available upon publication.}

\subsection{Models}
We consider two transformer \citep{vaswani2017attention} language model architectures: masked and autoregressive. For the MLM, we used the same architecture as BabyBERTa \citep{huebner-etal-2021-babyberta}, which is a small version of RoBERTa \citep{liu2019robertarobustlyoptimizedbert} and consists of 8 layers,
8 attention heads per layer, 256 hidden units, and an intermediate size of 1024. For the autoregressive language model, we adopted the GPT-2 architecture \citep{radford2019language} --- specifically, the smallest size of GPT-2, which consists of 12 layers, 12 attention heads per layer, and  768 hidden units. 

\subsection{Data} \label{subsec:data}
To better simulate the child acquisition process, we used the developmentally-plausible CHILDES \citep{whinney2000} corpus as provided by the BabyLM challenge (\url{https://babylm.github.io/}), where the data have been divided into train, dev, and test splits. We use a subset of this corpus that consists entirely of child-directed speech. This leaves approximately 3M sentences (around 10M tokens).

We create the following two data perturbations based on the \textsc{original} version for all train, dev, and test splits: \textsc{replace.word} and \textsc{shuffle.order}. Examples are shown in Table \ref{tab:data_summ}. Critically, the amount of training data remains the same under all three conditions, since neither perturbation changes the quantity of data. Thus, any differences seen across these conditions can be attributed to the type of data rather than the quantity.

\begin{table*}[t]
    \centering
    \begin{tabular}{ccc}
    \toprule
 \textbf{Type} & \textbf{Perturbation} & \textbf{Example} \\
      \midrule
    \rowcolor{mygrey}\textsc{original}   & N/A  &  you can \underline{eat} your supper when it's cooked. \\
    \rowcolor{myyellow}\textsc{replace.word} & co-occurrences
    &  you can \underline{eat} your \textbf{one} when it's \textbf{built}.\\
    \rowcolor{myblue}\textsc{shuffle.order}  & word order & can when cooked eat it's your you supper. \\
    
    \bottomrule
    \end{tabular}
    \caption{Summary of training perturbations and examples. Bolded words in \textsc{replace.word} are replacement words, while underlined words are the ones that remain unchanged from \textsc{original}.}
    \label{tab:data_summ}
\end{table*}

\paragraph{\textsc{replace.word}} In the \textsc{replace.word} perturbation, we remove the distributional co-occurrence information that is associated with a given verb. 
Nouns, adjectives, and adverbs that co-occur with the verb in the sentence are replaced with other words with the same part-of-speech (PoS), where the replacement words are sampled by frequency within the PoS to preserve the overall frequency distribution. For sentences with multiple verbs, the main verb remains unchanged, while the embedded verbs are sampled in the same way as other PoS categories. The resulting dataset retains the syntactic structure that is associated with a verb but has relevant co-occurrence information removed. 
Therefore, when we perform evaluations based on verb meaning, a learner that uses syntactic information to learn verbs should do well when trained with \textsc{replace.word}, while a learner that relies more on distributional information is expected to perform worse in the same situation. 

\paragraph{\textsc{shuffle.order}} In the \textsc{shuffle.order} perturbation, we removed the word order dependencies that existed in \textsc{original}. The intuition is that if models rely on syntactic information to learn verb meanings, ones that have been trained on \textsc{shuffle.order} will perform poorly in a targeted evaluation task for verb meanings. In comparison, if a model relies more on co-occurrence statistics, the model should succeed with this data, as co-occurrence information of words in the same sentence is preserved. We used the same SentenceRandomizer algorithm as in \citet{sinha-etal-2021-masked}.

Within \textsc{shuffle.order}, we considered two subtypes, each created under different shuffling conditions. The first subtype is \texttt{1gram}, where words are shuffled on the scale of unigrams. The other subtype is \texttt{np}, where we first used spaCy \citep{spacy2} to identify base noun phrases and treated them as a single unigram during shuffling.  
Therefore, \texttt{np} and \texttt{1gram} are only distinguished in cases with more complex NPs, which might be reasonably rare in child-directed speech. Since the number of NPs is an important cue for inferring the meaning of verbs, we expect models that are trained on \texttt{np} to have a better representation of verb meanings than those trained on the \texttt{1gram} perturbation.

\subsection{Training}

The training batch size and learning rate were determined via a hyperparameter search process. We found that for both architectures, models perform best under a learning rate of 1e-4. The optimal training batch size is 128 for RoBERTa and 256 for GPT-2.

We trained the two architectures on each of the three dataset versions for 5 epochs, and each training configuration is repeated with 5 different random seeds. All the following results are averaged across these 5 random initializations.

\subsection{Evaluation} \label{subsec:eval}

Depending on the model's architecture, we adopted different evaluation tasks for probing how well models encode verb meanings.

\paragraph{Masked Verb Prediction}
To evaluate our models trained under the RoBERTa architecture, we took inspiration from the Human Simulation Paradigm \citep{gillette1999human}, where adult participants were asked to guess verbs based on a range of cues. In our masked verb prediction (MVP) dataset, we mask out the verb in a sentence and evaluate the model on its ability to select the correct verb. Since we are interested in the effect of training perturbations on learned representations, the examples in this data set are selected from the \textsc{original} sequences, holding the evaluation set constant for all models. In addition, we restrict attention to sequences in which the original verb is tokenized as a single token. The resulting test set contains 109,984 sentences. For each sentence, we replaced the main verb with the mask token \texttt{<mask>} and then passed the modified sentence through the model. The output logits at the masked position are extracted, and a prediction is considered correct if the top-ranked token corresponds to the original verb --- e.g., if \textit{tell} is the top-ranked token when the model is given the masked version of (\ref{ex:tell}):
\ex[aboveexskip=4pt, belowexskip=4pt]<maskex> \label{ex:tell}
Original: Can you tell me more?\\
Masked: Can you \texttt{<mask>} me more?\\
Correct: \textit{tell}
\xe

\paragraph{Minimal Pair Judgment}
For our models trained under the autoregressive GPT-2 architecture, the model cannot be evaluated on the task of MVP. Instead, we probe models' knowledge of verb meanings through the probability they assign to a sentence $S$ that is acceptable in comparison to another unacceptable sentence $S'$ that differs from $S$ only by a verb. 
Similar to MVP, we curate the evaluation set based on the test split of \textsc{original}. Since the acceptability of shorter sentences varies less under different verb choices (e.g., simple imperatives such as \textit{[verb] it} allow almost any verb and often occur in child-directed speech), we restrict our evaluation set only to sentences that are more than ten words long. First, we used spaCy to identify the main verb. Second, we sampled five other verbs within the same frequency bin as the original verb. This is achieved by first ranking all verbs in the test corpus by descending frequency and then binning them into four groups based on their log-rank values. For each sentence, substituting the main verb with the replacement verb results in five alternative sentences, which correspond to five total minimal pairs. Notice that our approach does not guarantee that all alternative sentences are strictly unacceptable. However, it is reasonable to assume that the alternative sentences should be less well-formed than the original sentence given the sentential context. In addition, the selection of five verbs instead of one also serves to minimize the effect that a specific replacement verb can have on the accuracy results.
\pex[aboveexskip=5pt, belowexskip=5pt]<mpex>
\a if you don't \textbf{want} a new mother you better be good.

\a \# if you don't \textbf{\{play, push, give, stick, listen\}} a new mother you better be good.
\xe
For each minimal pair test item $\langle S,S'\rangle$, we consider models to perform correctly if they assign a higher probability $S'$ to the original sentence than to the unacceptable sentence $S'$:
\begin{equation} \label{eq:psych}
    P(S) > P(S')
\end{equation}

\section{Experiment 1: LLMs rely more on syntax than co-occurrence to learn verbs} \label{sec:exp1}

\begin{figure}[t]
    \centering
    \includegraphics[width=\linewidth]{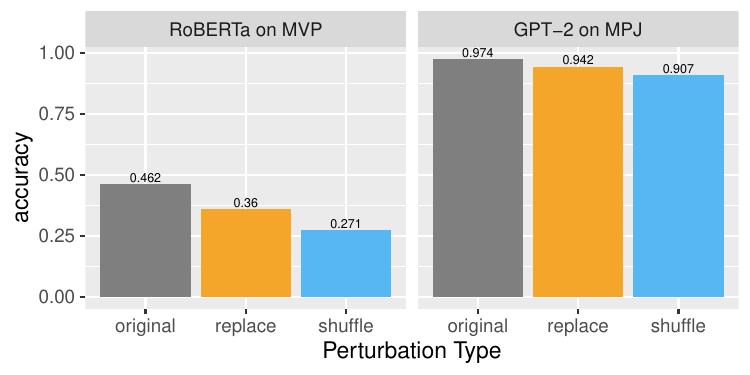}
    \caption{Accuracy of RoBERTa and GPT-2 on the Masked Verb Prediction (MVP) and Minimal Pair Judgment (MPJ) task, respectively, by perturbation type. For RoBERTa under MVP (left panel), chance performance is $\frac{1}{30000}$ given a vocabulary size of 30000. For GPT-2 under MPJ (right panel), chance performance is 50\%.}
    \label{fig:modeloverall}
\end{figure}

The accuracy of RoBERTa and GPT-2 on the Masked Verb Prediction and Minimal Pair Judgment tasks is shown in Figure \ref{fig:modeloverall}.
In the \textsc{original} case, both architectures perform far above chance, indicating that our general setup is sufficient to enable models to learn a substantial amount about verb meaning. Turning to the perturbations, we find that both \textsc{replace.word} and \textsc{shuffle.order} have lower performance than \textsc{original}, suggesting that both syntactic information and word co-occurrences are helpful in learning verb meanings. More interestingly, across both architectures, models trained on the \textsc{shuffle.order} perturbation have the lowest accuracy, with \textsc{replace.word} being relatively higher. The ranking \textsc{original} > \textsc{replace.word} > \textsc{shuffle.order} suggests that removing syntactic information has a larger effect on the model's verb knowledge than removing co-occurrence information. We take this pattern to be the first piece of evidence showing the important role of syntax for transformer language models to acquire verb meaning.

\paragraph{Mental vs.\ Physical Verbs}Recall from Section~\ref{sec:synboo} that the identification of mental verbs during the Human Simulation Paradigm benefits more from syntactic cues than physical verbs. Therefore, if transformer language models do employ syntactic information in a similar way to human syntactic bootstrapping, we would not only observe the \textsc{original} > \textsc{replace.word} > \textsc{shuffle.order} ranking but also an interaction between verb type and perturbation type. Specifically, the performance degradation from \textsc{original} to \textsc{shuffle.order} should be larger for mental verbs than for physical verbs.
By our hypothesis, the degradation from \textsc{original} to \textsc{replace.word} should also give us a sense of how co-occurrence information affects verb learning. While we have no strong expectation a priori how mental vs.\ physical verbs might differ under the loss of co-occurrence, one possibility is that physical verbs might suffer more in this case, given the stronger reliance on structure for mental verbs.

\begin{figure}[t]
    \centering
    \includegraphics[width=1\linewidth]{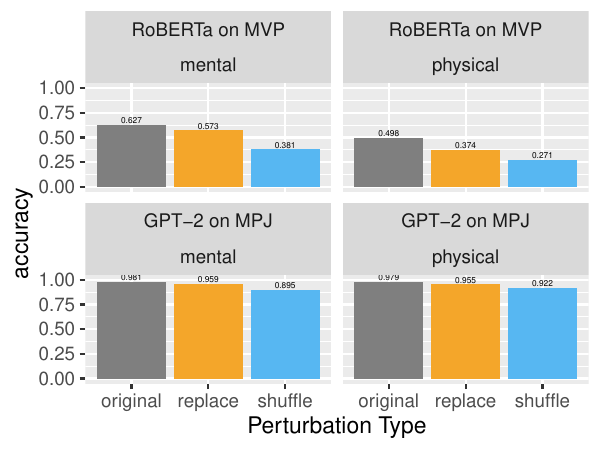}
    \caption{Accuracy on a selected set of verbs by verb type (left vs.\ right panel), model architecture (top vs.\ bottom panel), and perturbation type (x-axis).}
    \label{fig:verbmentalphysical}
\end{figure}

We now unpack the results in Figure \ref{fig:modeloverall} to examine whether this pattern holds. We used the exact same set of mental and physical verbs as \citet{gillette1999human}. 
\pex[aboveexskip=5pt, belowexskip=5pt]<mentalphysical>
\a Physical: \textit{take}, \textit{say}, \textit{come}, \textit{play}, \textit{push}, \textit{sit}, \textit{pull}, \textit{eat}, \textit{make}, \textit{call}, \textit{catch}, \textit{put}, \textit{find}
\a Mental: \textit{see}, \textit{look}, \textit{want}, \textit{know}, \textit{like}, \textit{think}
\xe 
Results for MVP and MPJ evaluations with sentences involving only these verbs are shown in Figure \ref{fig:verbmentalphysical}. 

First, the \textsc{original} > \textsc{replace.word} > \textsc{shuffle.order} ranking continues to hold for this smaller set of verbs across both architectures. Second, recall that if mental verbs rely more on syntax than physical verbs, we would expect the performance degradation from \textsc{original} to \textsc{shuffle.order} to be bigger for mental verbs. This is exactly what we observe for both RoBERTa and GPT-2. For example, while the accuracy for physical verbs drops only 5 points from \textsc{original} to \textsc{shuffle.order} in GPT-2, the corresponding degradation for mental verbs is 9 points. To test for statistical significance, we fit a logistic mixed-effect model using the \texttt{lme4} package in R with the following formula for each model architecture: \texttt{correct \textasciitilde{} perturb\_type * verb\_type + (1|correct\_answer\_id)}, where \texttt{correct} is a binary variable indicating whether the model performs correctly on a specific test item.
The \textsc{shuffle.order} × physical interaction is positive for both RoBERTa ($\beta = 0.02, p<0.001$) and GPT-2 ($\beta = 0.03, p<0.001$), which means that the representation of mental verbs indeed suffers more than the representation of physical verbs when syntactic information is taken out from the training data.

On the other hand, the \textsc{replace.word} × physical interaction term is significant for RoBERTa ($\beta=-0.07, p<0.001$) but not for GPT-2 ($\beta = -1.76e-03, p=0.282$). This means that in RoBERTa, the representation of physical verbs suffers more than the representation of mental verbs from losing co-occurrence information, while the effect of such removal does not differ significantly between verb types in GPT-2. While it is not entirely clear to us why there might be a difference between RoBERTa and GPT-2 in the \textsc{replace.word} × physical interaction, it might stem from differences in the task that each architecture performs (i.e., masked language modeling vs.\ autoregressive next-word prediction). We leave the exact mechanism under which this might happen for future research.
Overall, the LLM results are highly consistent with the findings from the Human Simulation Paradigm in \citet{gillette1999human}. Like humans, LLMs rely more on syntactic structure than co-occurrence information in learning verb meanings, and such dependence on syntax is stronger for mental verbs.

\section{Experiment 2: LLMs rely more on co-occurrence than syntax to learn nouns} \label{sec:exp2}

In Experiment 1, we showed the important role of syntactic information during LLM verb learning. However, there is the possibility that the \textsc{replace.word} perturbation simply removes less information in some absolute sense from the data than the \textsc{shuffle.order} perturbation, which could explain the results we have observed but in a way that is driven by amount of information rather than a distinction between syntax and co-occurrence. We address this concern by focusing on the distinction between noun and verb learning. Under the syntactic bootstrapping hypothesis, a small set of nouns is learned first, prior to the point at which syntactic distribution could be attended to, which paves the way for subsequent syntax and verb learning. In addition, results in the Human Simulation Paradigm suggest that compared to guessing verbs, the identification of nouns is much easier given situational cues. If this holds for LLMs as well, the learned representation of nouns when co-occurrence information is removed will be worse than that when syntactic information is removed.
To test this prediction, we conduct a variant of our first experiment, focusing instead on nouns. All experimental setups are the same as Experiment 1 except for the \textsc{replace.word} training perturbation and the targeted evaluation set on nouns.

\paragraph{Training Data} To examine the role of co-occurrence information on noun learning, we create a new \textsc{replace.word} perturbation where the verbs, adjectives, and adverbs that co-occur with a noun in a sentence are replaced with other words in the same PoS. For sentences with multiple nouns, we randomly choose a noun that will remain unchanged and replace the others. This perturbation is the exact parallel of the \textsc{replace.word} perturbation for verbs, where we replaced the nouns, adjectives, and adverbs that co-occur with a verb. Examples under this perturbation are shown in Table \ref{tab:exp2summary} along with \textsc{original} and \textsc{shuffle.order} to demonstrate the difference.

\begin{table*}[t]
    \centering
    \begin{tabular}{ccc}
    \toprule
 \textbf{Type} & \textbf{Perturbation} &\textbf{Example} \\
      \midrule
    \rowcolor{mygrey}\textsc{original}     & N/A &  you can \underline{eat} your \underline{supper} when it's cooked. \\
 
    \rowcolor{myyellow}\textsc{replace.word} (Exp1) & 
    verb co-occurrences &  you can \underline{eat} your \textbf{one} when it's \textbf{built}.\\
    \rowcolor{myyellow}\textsc{replace.word} (Exp2) & noun co-occurrences
    &  you can \textbf{say} your \underline{supper} when it's \textbf{trapped}.\\

       \rowcolor{myblue}\textsc{shuffle.order} & word order  & can when cooked eat it's your you supper. \\
    \bottomrule
    \end{tabular}
    \caption{Replication of Table \ref{tab:data_summ} with additional row summarizing the data perturbation used in Experiment 2.}
    \label{tab:exp2summary}
\end{table*}

\paragraph{Evaluation}
The evaluation of noun representation closely follows that of verbs: \textbf{Masked Noun Prediction}, where RoBERTa models were evaluated on whether they can correctly predict the masked noun in a sentence, and \textbf{Minimal Pair Judgment}, where the words that differ between an acceptable sentence and an unacceptable one are nouns, and the probability of each sentence was extracted from GPT-2. Similar to the case of verbs, the replacement nouns were also sampled from the same frequency bin as the original noun. See below for examples:
\ex[aboveexskip=5pt, belowexskip=5pt]<mnp>
Masked Noun Prediction\\
Original: can you sing happy birthday?\\
Masked: can you sing happy \texttt{<mask>}?\\
Correct: \textit{birthday}
\xe 

\pex[aboveexskip=5pt, belowexskip=5pt]<mpnoun>
Minimal Pair Judgment:
\a I'll buy you one down the \textbf{beach} for ten cents.
\a \# I'll buy you one down the \textbf{\{wave, motor, bite, turkey\}} for ten cents.
\xe 

\begin{figure}[ht]
    \centering
    \includegraphics[width=1\linewidth]{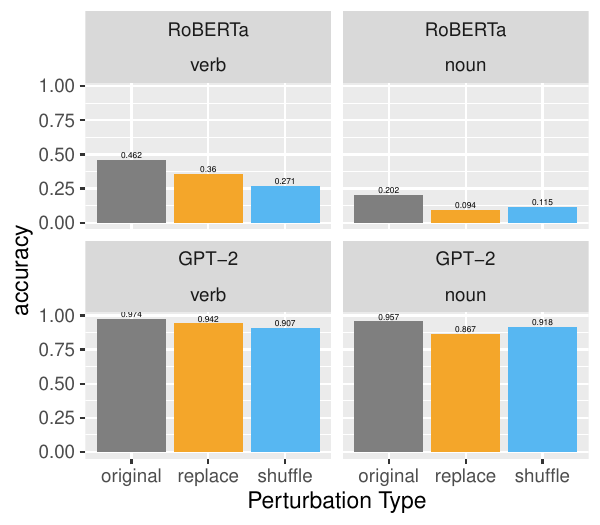}
    \caption{Accuracy by part of speech (left vs.\ right panel, where left panel is the same as Figure \ref{fig:modeloverall}), model architecture (top vs.\ bottom panel), and perturbation type (x-axis). Consistent with Experiment 1, RoBERTa models are evaluated under the Masked Noun Prediction task as in (\getref{mnp}), while GPT-2 models are evaluated under the Minimal Pair Judgment task as exemplified in (\getref{mpnoun}).}
    \label{fig:verbnounaccuracy}
\end{figure}

\paragraph{Results}  As shown in Figure \ref{fig:verbnounaccuracy}, the accuracy pattern for nouns is qualitatively different from that of verbs. For verbs, as shown in the left panel of Figure \ref{fig:verbnounaccuracy} (which is a replication of Figure \ref{fig:modeloverall}), models under both architectures exhibit the ranking \textsc{original} > \textsc{replace.word} > \textsc{shuffle.order}. However, when it comes to the representation of nouns, \textsc{replace.word} becomes the perturbation with the lowest accuracy: \textsc{original} > \textsc{shuffle.order} > \textsc{replace.word}. All contrasts are statistically significant ($p<0.001$) under a mixed-effect model with perturbation type as the main effect predictor and correct answer ID as the random intercept.
In other words, the learning of nouns is more dependent on co-occurrence information as compared to syntactic structure, while the learning of verbs shows the reverse pattern. The stark contrast between verb and noun performance further aligns with human results and emphasizes how the same training data perturbation can affect the learning of different linguistic elements in different ways. 

\begin{figure}[t]
    \centering
    \includegraphics[width=1\linewidth]{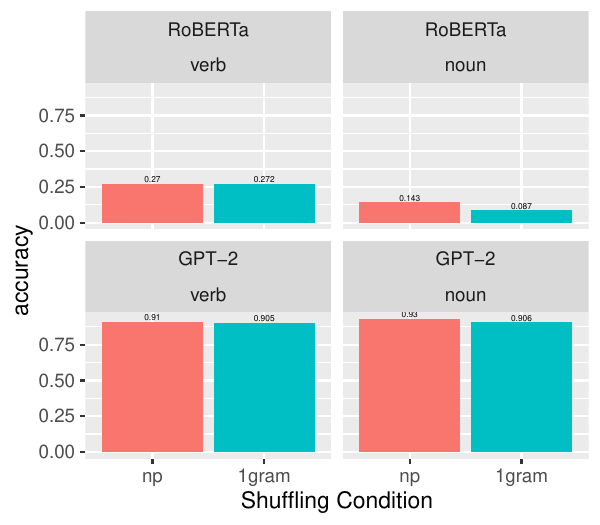}
    \caption{Accuracy under \textsc{shuffle.order} by two shuffling conditions (\texttt{np} vs.\ \texttt{1gram}), model architectures, and part-of-speech.}
    \label{fig:shuffleaccuracy}
\end{figure}

As a side note, the prediction that maintaining noun phrases during shuffling will help verb acquisition was not borne out. As shown in Figure~\ref{fig:shuffleaccuracy}, \texttt{np} has a similar accuracy as \texttt{1gram} on the left panel, which means that the size of the shuffling condition does not play a significant role in the acquisition of verb meaning across both model architectures. In other words, maintaining the boundary of noun phrases does not help models learn better verb representations. However, such information is helpful for noun learning (\texttt{np} > \texttt{1gram} in the right panel), which is reasonable given that the other parts of the noun phrase can be predictive of the noun itself.

\section{Discussion} \label{sec:discussion}
In this paper, we examined whether LLMs exhibit syntactic bootstrapping by training the RoBERTa and GPT-2 architectures on perturbed variants of the CHILDES corpus: \textsc{shuffle.order}, which ablates syntactic information, and \textsc{replace.word}, which removes the co-occurrence between verbs and the other content words that occur in the same sentence. Results showed a strong reliance on syntax for verb meaning: models' accuracy on a targeted verb meaning evaluation set decreases the most for \textsc{shuffle.order}, as compared to \textsc{original}. What's more, both model architectures exhibit the same mental vs.\ physical distinction as humans, where mental verbs rely more on syntactic structure than physical ones. The consistency across both RoBERTa and GPT-2 shows that the reliance on syntax for verbs is not an artefact of some specific model architecture. It might be a general mechanism that holds across training objectives. This finding is strengthened in Experiment 2, where we showed that the acquisition of nouns exhibits the reverse pattern --- co-occurrence information matters more than syntax.

\paragraph{Distribution as a proxy for meaning} Our evaluation method uses distribution as a proxy for meaning. In other words, we assumed that models' ability to accurately predict word identity (as in Masked Noun/Verb Prediction for RoBERTa) or to (dis)prefer certain words (as in Minimal Pair Judgment for GPT-2) in a given sentence frame reflects their representation of verb meanings. This assumption is supported by the Distributional Hypothesis \citep{harris_distributional_1954, firth_synopsis_1957}, which holds that word distribution predicts word meaning. This idea also underlies modern NLP developments such as word2vec, where the model is trained to predict a word using its neighboring contexts. The resulting word embeddings successfully encode a wide range of syntactic and semantic relations such as gender, tense, and number \citep{mikolov2013efficientestimationwordrepresentations}. Compared to word2vec, our evaluation method lies on the behavioral side --- we treat LLMs' predictions and the probabilities that they assign to sentences as a reflection of their underlying encodings of meaning, instead of focusing on the geometric properties of learned word embeddings. This design followed the setup of the Human Simulation Paradigm \citep{gillette1999human}, which is also on the behavioral level. This paradigm draws conclusions about people's ability to infer word meaning based on whether they can identify words from 
distributional cues, including syntactic frames and word co-occurrences. Given similar setups in prior work, we believe that distribution is a reliable proxy for meaning and is a helpful way to assess meaning representations within LLMs.

\paragraph{Implications for developmental work}

Overall, our results strengthen findings from the literature on the Human Simulation Paradigm and the syntactic bootstrapping hypothesis: even for a weakly-biased learner that does not have strong predispositions tailored to the structure of language, syntactic structure still plays a greater role than co-occurrence information in determining learned verb representations. This means that the statistical regularities that exist between verbs and syntactic structure must be sufficiently stable such that they cannot be ignored by any learner with an effective learning procedure that approximates the underlying distribution of language.
Furthermore, the same types of information that facilitate human verb learning also support learning in machines. 

To be clear, the current results provide evidence for an alignment between LLMs and humans in the types of information that they leverage for learning word meanings, but they do not speak directly to questions of whether LLMs and humans use the same learning mechanisms to exploit this information.
Our finding highlights the important role of syntax during the LLM learning process: in order to arrive at an accurate representation of verb meaning and hence language, LLMs must somehow rely on syntactic structure. While the exact way in which such regularities factor into the learning process is beyond the scope of the current paper, our approach of transferring syntactic bootstrapping to LLMs identified one direction that can be unpacked on the circuit and mechanistic level. Investigating the exact mechanisms could shed light on the extent to which LLMs mirror human language acquisition and the plausibility of LLMs as cognitive models of the human mind.

\paragraph{LLMs as a tool for testing developmental hypotheses on a larger scale} From a broader point of view, our results also demonstrated how the paradigm of training data manipulation can facilitate theoretical and experimental work in language acquisition. Since our data was taken from a developmentally plausible corpus, the results from models trained under different perturbations can reveal the kinds of cues that are available in the naturalistic input and their relevant contributions to the learning process. These manipulations are often not possible with actual children and need to be carefully designed and controlled in child language experiments. In comparison, our methodology not only allows for systematic selection of linguistic input but also simulates the entire course of acquisition, which extends the time scale on which the effects of a certain linguistic element can be observed. While the current paper only focuses on verb learning, future studies can generalize this methodology to other aspects of the acquisition process (e.g., adjective learning) and study the degree to which humans and LLMs differ.

\paragraph{Other cues to word learning}
One limitation of the current paper is that it only focused on the relative strength of two cues for verb and noun learning: syntactic structure and co-occurrences of content words. Other linguistic signals might also play a role in this process, including cues based on function words and sentence length. It would be interesting to isolate the strength of these components and see what the most crucial driving factors are for verb learning, which we leave for future research. In addition, our paper only examined the linguistic cues to word learning, which in reality is a multimodal process grounded in social interactions \citep{bloom2002children}. While existing work has begun to explore whether neural networks can learn object names via visual grounding \citep{vong2024grounded} and the benefit of multimodality for word learning \citep{zhuang-etal-2024-visual}, much remains unknown about how these information sources interact for all aspects of language. Extending our paradigm to other modalities could take us one step closer to modeling the actual environment in which children learn language.

\section{Conclusion} \label{sec:conclusion}

In this paper, we argue that syntactic bootstrapping, a hypothesis of human verb acquisition with strong empirical support, can be used to better understand acquisition in LLMs. We found that LLMs rely on syntactic structure to learn verb meanings in human-like ways: compared to syntax, removing co-occurrence information causes a smaller degradation to the learned verb representations. Although our work successfully reveals the human-machine alignment in verb acquisition, we did not examine the full learning trajectory. Instead, our analysis focused on representations after training had completed, leaving open the question of how such representations are learned over training steps. As argued by \citet{gillette1999human}, verb learning requires some representation of syntax, which could not occur until a small set of concrete nouns is learned. Therefore, future work could focus on examining the time points at which verb learning and noun learning differ, which will offer a more fine-grained view of how and when alignment with human learning arises.

\section*{Acknowledgements}

We thank the Computational Linguistics at Yale lab (CLAY) and the Yale Linguistics Department for helpful comments and discussion. We thank the Yale Center for Research Computing and Wu Tsai Institute for providing computational resources. Any errors are our own.

\bibliography{anthology, my}
\bibliographystyle{acl_natbib}

\iftaclpubformat

\onecolumn

\appendix

\fi

\end{document}